\documentclass[10pt,twocolumn,letterpaper]{article}

\usepackage{iccv}
\usepackage{times}
\usepackage{epsfig}
\usepackage{graphicx}
\usepackage{amsmath}
\usepackage{amssymb}

\usepackage{booktabs}
\usepackage{multirow}
\usepackage{color}
\usepackage{textcomp}
\usepackage{float}
\usepackage{authblk}

\usepackage[breaklinks=true,bookmarks=false]{hyperref}

\iccvfinalcopy 

\def\iccvPaperID{1850} 

\ificcvfinal\fi

\begin{document}

\title{Moving Indoor: Unsupervised Video Depth Learning in Challenging Environments}

\author[1,\footnote{}]{Junsheng Zhou}
\author[2]{Yuwang Wang}
\author[1]{Kaihuai Qin}
\author[2]{Wenjun Zeng}
\affil[1]{Tsinghua University, Beijing, China \authorcr
{\tt\small zhoujs17@mails.tsinghua.edu.cn, qkh-dcs@mail.tsinghua.edu.cn}}
\affil[2]{Microsoft Research, Beijing, China \authorcr
{\tt\small {yuwwan, wezeng}@microsoft.com}}


\maketitle

\footnotetext[1]{Work done as an intern at MSRA.}
\ificcvfinal\thispagestyle{empty}\fi

\begin{abstract}
Recently unsupervised learning of depth from videos has made remarkable progress and the results are comparable to fully supervised methods in outdoor scenes like KITTI. However, there still exist great challenges when directly applying this technology in indoor environments, e.g., large areas of non-texture regions like white wall, more complex ego-motion of handheld camera, transparent glasses and shiny objects. To overcome these problems, we propose a new optical-flow based training paradigm which reduces the difficulty of unsupervised learning by providing a clearer training target and handles the non-texture regions. Our experimental evaluation demonstrates that the result of our method is comparable to fully supervised methods on the NYU Depth V2 benchmark. To the best of our knowledge, this is the first quantitative result of purely unsupervised learning method reported on indoor datasets.
\end{abstract}

\section{Introduction}

Reconstructing the structures of scenes from RGB images has long been a popular research topic. Depth estimation is an important step to reconstruct the scenes. It is easy for human to perceive the depth of scenes from a RGB image since we have prior knowledge about the sizes of common objects and their layouts, but it is difficult for computer to estimate accurate depth map from a single image due to scale ambiguity. Classical methods like Structure-from-Motion and Stereo Matching~\cite{furukawa2015multi,hartley2003multiple} were proposed and achieved plausible results. In recent years, due to the surge of deep learning, convolutional neural network was introduced to predict depth from monocular RGB images for its powerful capacity of feature extraction,which leads to great progress in this field. These methods~\cite{liu2014discrete,eigen2014depth,eigen2015predicting}, treat the neural networks as blackbox with strong fitting ability and use the collected ground-truth annotation to supervise the training. However, these fully supervised methods are limited by the huge demand of training samples. 

Recently unsupervised depth learning~\cite{zhou2017unsupervised,godard2017unsupervised,yin2018geonet} has been proposed and attracted more and more interests. This method is similar to traditional Structure-from-Motion that leverages the disparity information contained in videos to supervise the networks' training. The key idea is novel view synthesis based on simultaneously estimated depth of scene and ego-motion of camera. The appearance difference between the synthesized view and real view is used as supervisory signal for the entire training pipeline. These unsupervised methods do not need ground-truth annotation and achieve remarkable results on driving scenes like KITTI and Cityscapes. 

However, there still exist great challenges when directly applying this technology in indoor environments. In our experiments, we observed that the same model and same training setting, which is able to achieve state-of-the-art performance on KITTI, soon collapse when training on indoor datasets NYU V2\cite{silberman2012indoor} and Scannet~\cite{dai2017scannet}. The reason is that indoor environments are more complicated than urban driving scenes. The main problems can be summarized below:

1) Large areas of non-texture regions. Unlike fully supervised methods in which each pixel has ground-truth supervision, the supervisory signal of unsupervised learning only comes from the appearance difference between images themselves. Non-texture regions seriously hinder the training since in these regions the photometric loss is always close to zero. However we observed that there are considerable amount of images which have more than $50\%$ non-texture areas in indoor datasets. White wall and carpets are fairly common non-texture objects.

2) More complex ego-motion of handheld camera. Generally indoor datasets are collected by handheld cameras, which means that the ego-motion of consecutive frames is more complex than driving scenes where the cars are mostly just moving forward. Especially, we can not infer the depth of scene theoretically from sequences of pure rotation. The existence of large amount of training samples with pure rotation will overwhelm the entire training process. 

These reveal that the existing training pipeline needs to be revamped to be applied in more general scenes. In this paper, we propose a new optical-flow based training paradigm, which focuses on the most important part, i.e., the supervisory signal of unsupervised depth learning. This new pipeline uses the optical flow results generated by a flow estimation network as supervision and it is easier for the training to converge. The key component is a specially designed network which is responsible for estimating the optical flow between consecutive frames in a sparse-to-dense propagation manner. This unsupervised flow network is able to handle the non-texture regions and produce plausible optical flow results. Then this network can be used as a teacher simultaneously training the DepthNet and the PoseNet. We also improve the existing PoseNet and make it easier to learn the complex ego-motion of handheld camera.

Our evaluation on NYU Depth V2 benchmark demonstrates that the result of our method is comparable to fully supervised methods. 

\section{Related Work}

\textbf{Supervised Depth Estimation}\quad Estimating the depth from a single image has been long studied. Recently due to the success of deep learning~\cite{li2019_internet,li2019_cmt}, many networks for depth estimation have been proposed~\cite{eigen2014depth,eigen2015predicting,li2017two,liu2014discrete,fu2018deep,laina2016deeper}. Eigen et al.~\cite{eigen2014depth} applied multi-scale networks which first estimate coarse depth by a coarse-scale network and refine it by another network. CRF is also introduced to this task and used as a postprocessing module in a model~\cite{xu2017multi}. Fu et al.~\cite{fu2018deep} treated the depth estimation as a classification problem instead of a regression problem. All these methods have powerful capacity and achieve very good performance on both indoor datasets like NYU V2~\cite{silberman2012indoor}, Scannet~\cite{dai2017scannet} and outdoor datasets like KITTI~\cite{geiger2012we}, Make3D~\cite{saxena2006learning,saxena2009make3d}. However these methods rely on large-scale datasets with depth labels.

\textbf{Unsupervised Depth Learning}\quad To get rid of the need of ground-truth depth annotation, unsupervised depth learning methods have been proposed. These methods leverage either stereo images~\cite{godard2017unsupervised,yang2018every} or videos~\cite{zhou2017unsupervised,yin2018geonet} as training data. Godard et al.~\cite{godard2017unsupervised} first proposed to use the left-right consistency of stereo images to train a depth estimation network. Zhou et al.~\cite{zhou2017unsupervised} applied two networks which jointly estimate the depth and ego-motion of camera to learn the depth from videos. Wang et al.~\cite{wang2018learning} discarded the pose network and directly computed the pose by visual odometry method. Casser et al.~\cite{casser2019struct2depth} leveraged the additional instance segmentation masks to model the dynamic objects. These methods have achieved tremendous success in outdoor scenes like KITTI~\cite{geiger2012we} and CityScapes~\cite{cordts2016cityscapes}. However, only some of these works demonstrated the sample predicted results of indoor scenes using the networks trained on KITTI, and there is no quantitative result reported on typical indoor datasets. In our experiments, we also faced great challenges when directly using the previous methods to train on indoor scenes.

\textbf{Unsupervised Optical Flow Learning}\quad Based on the same photometric supervisory signal as unsupervised depth learning, unsupervised optical flow learning methods have also been proposed. Yu et al.~\cite{jason2016back} and Ren et al.~\cite{ren2017unsupervised} proposed FlowNet-based~\cite{ilg2017flownet} architecture for unsupervised optical flow learning. Meister et al.~\cite{meister2018unflow} proposed a Bidirectional Census Loss to handle occlusion/disocclusion. Although these methods perform well on synthetic datasets, non-texture regions, dynamic objects and occlusion are still intractable problems. Our new flow network is different from previous architecture, which leverages the sparse flow seeds generated by traditional feature matching methods and progressively propagates them to the entire image. Non-texture regions are handled well by this mean.

\begin{figure*}
\begin{center}
  \includegraphics[width=1\linewidth]{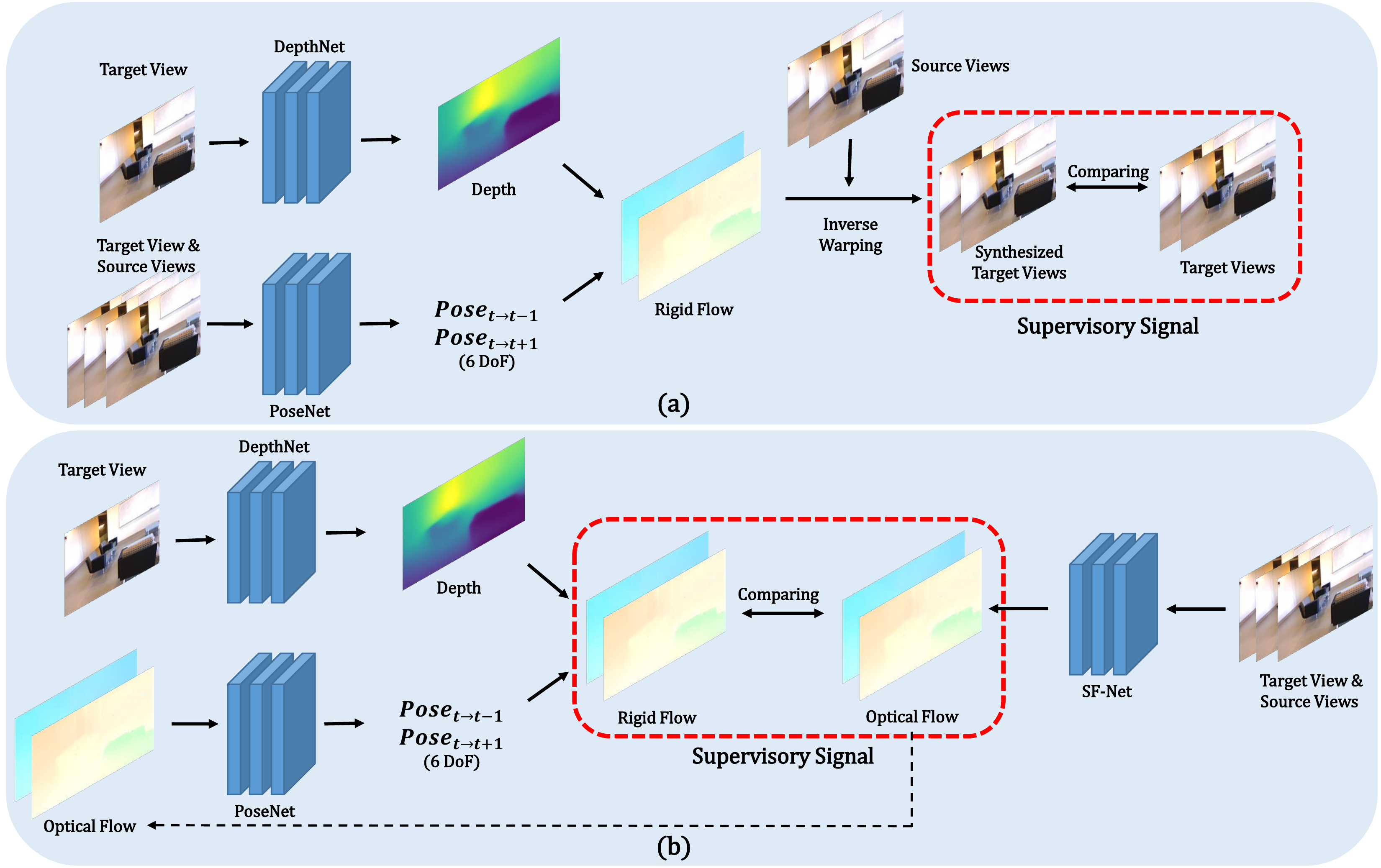}
\end{center}
   \vspace{-3mm}
   \caption{Overview of previous pipeline (a) and our pipeline (b). The supervisory signal of previous pipeline is based on appearance matching of two images, which is unstable and suffers from non-texture regions. We use the optical flow results generated by our flow estimation network SF-Net as supervision and modify the input of PoseNet.}
\label{fig:overview}
\vspace{-2mm}
\end{figure*}

\section{Approach}

\subsection{Overall Pipeline}

As shown in Figure~\ref{fig:overview} (a), previous unsupervised depth learning pipeline generally consists of two modules: DepthNet and PoseNet. During training, both networks simultaneously estimate the depth of the scene and the ego-motion of the camera. Once the depth map of RGB image is estimated, we can backproject the pixels on image plane to 3D coordinates with known camera intrinsic. Then with the estimated motion the 3D point cloud can be transformed to another view. This view transformation can be formulated as below:
\begin{equation}
    p_{s\xrightarrow{}t} = KT_{t\xrightarrow{}s}D_t(p_t)K^{-1}p_{t}
    \label{equ:flow}
\end{equation}
where $K$ denotes the camera intrinsic, $T_{t\xrightarrow{}s}$ denotes the estimated transformation matrix from view $t$ to view $s$, $D_t$ denotes the estimated depth, $p_t$ and $p_{s\xrightarrow{}t}$ denote the homogeneous coordinates of a pixel in view $t$ and view $s$ respectively. Then we can obtain the 2D rigid flow from view $t$ to view $s$:

\begin{equation}
    f_{t\xrightarrow{}s}(p_t) = p_{s\xrightarrow{}t} - p_t
\end{equation}
Once the rigid flow between two views is estimated, we can synthesize the image $I_{s\xrightarrow{}t}$ by differentiable inverse warping~\cite{jaderberg2015spatial} from source view. The appearance difference between the synthesized image $I_{s\xrightarrow{}t}$ and the real image $I_t$ is used as supervisory signal for the entire pipeline.

As shown in the red frame in Figure~\ref{fig:overview} (a), the photometric loss computed based on the synthesized image and the real image supervises the training of both networks. This is equivalent to finding the best match point in source view for each pixel in target view, which is similar to Stereo Matching. Once the rigid flow is predicted perfectly , the synthesized image matches the real image completely (if there is no occlusion and dynamic object). However, this appearance-based supervisory signal is indirect and susceptible to non-texture regions. More specifically, for each pixel there is no explicit target position that it should match in another view. The optimization target of this proxy objective is just to minimize the appearance difference between two images. This paradigm is difficult to work in indoor environments since there is a lot of non-texture stuff in our daily scenes, where the appearance difference is always close to zero, which can not provide valid and strong signal for the training.

To overcome the problem, we first look back to the entire pipeline. It can be separated into two stages: 1) composing the rigid flow based on jointly estimated depth and pose. 2) using the rigid flow to synthesize novel image and computing the loss (corresponding to the left and right parts in Figure~\ref{fig:overview} (a)). The objective function can be briefly written as:

\begin{equation}
    L = |{I_t - I_{s\xrightarrow{}t} } |
\end{equation}
where $I_t$ denotes target view image and $I_{s\xrightarrow{}t}$ denotes synthesized image. 

Our main contribution is that instead of using this indirect proxy supervision, we provide an explicit optical flow target to supervise the estimated rigid flow as shown in Figure~\ref{fig:overview} (b). The optical flow target is obtained by a sparse-to-dense flow estimation network (named hereafter as SF-Net), which will be elaborated on in the next subsection. Therefore the objective function is modified as:

\begin{equation}
    L = |f_{t\xrightarrow{}s}(p_t) - f'_{t\xrightarrow{}s}(p_t) |
\end{equation}
where $f'_{t\xrightarrow{}s}(p_t)$ denotes the optical flow result from SF-Net. This modification is significant since it transforms the unsupervised learning to ``fully supervised'' learning and reduces the difficulty of training. The words ``fully supervised'' here only mean more explicit supervisory signal while this signal is still obtained unsupervisedly. Photometric loss is not able to penalize the incorrect prediction in non-texture regions, i.e., the supervisory signal fails in those areas (common in indoor scenes). By contrast, the optical flow generated by SF-Net provides unique target for each pixel, which is a very strong supervisory signal. Our experiments show that this new pipeline is able to handle the indoor environments with large areas of non-texture regions and achieve plausible result.

\begin{figure}
\begin{center}
  \includegraphics[width=1\linewidth]{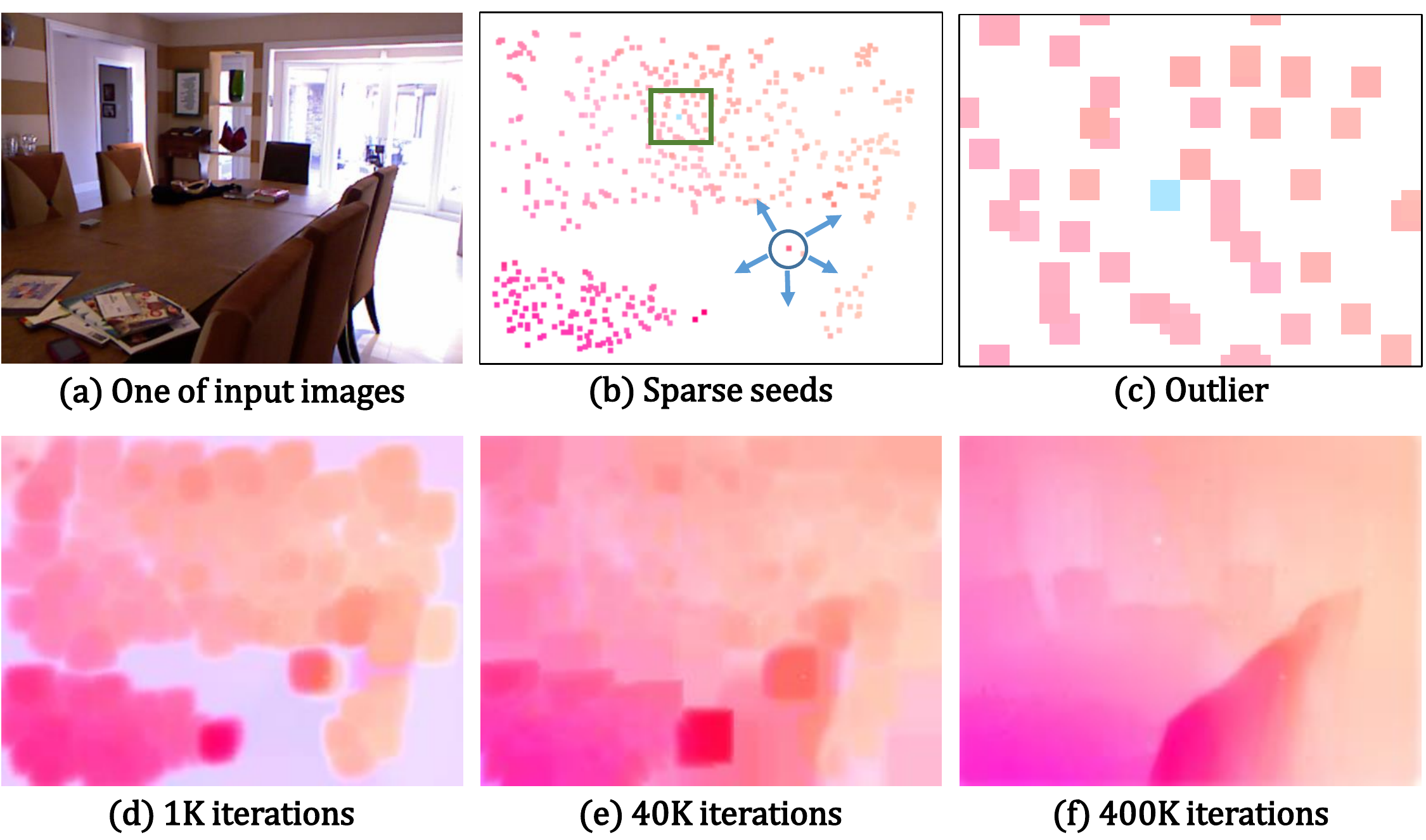}
\end{center}
   \vspace{-3mm}
   \caption{Illustration of our SF-Net. (a) One of input images. (b) Sparse seeds generated by SURF~\cite{bay2006surf}. The sizes of sparse points have been enlarged for better visualization. (c) The green frame in (b), which represents an outlier (blue point) in sparse seeds. (d)(e)(f) Visualization of a training sample at different stages. Our SF-Net works in a sparse-to-dense propagation manner which progressively propagates the sparse flow from textured regions to non-texture regions. The negative effects induced by outliers are also suppressed during training.}
\label{fig:flownet}
\vspace{-5mm}
\end{figure}

\subsection{SF-Net}

The core of this new training paradigm is the part of SF-Net. Before the introduction of the SF-Net, we first explain the principle of unsupervised optical flow learning. The key idea is novel view synthesis, which is the same as unsupervised depth learning. The network predicts a flow map for image $I_t$ in target view and use this dense flow to synthesize the image $I_{s\xrightarrow{}t}$ by differentiable inverse warping~\cite{jaderberg2015spatial} from source view. Then the appearance difference between the synthesized image $I_{s\xrightarrow{}t}$ and the real target image $I_t$ is used as supervisory signal for the training. 

However, since both unsupervised depth learning and optical flow learning leverage the same proxy supervisory signal, they also suffer from the same problems. Previous methods usually impose smoothness constraint on the flow prediction. The reason is that it is easy for the network to produce correct result in textured region like corner and border, we expect this correct prediction to guide its neighborhood where the prediction is incorrect. But in indoor environments where non-texture regions cover large areas, the correct prediction may be overwhelmed by the incorrect predictions on the contrary.

In order to overcome this problem, we propose an active propagation approach. Instead of only using the weak smoothness constraint, we actively propagate the sparse initial seeds at textured regions to the entire image as shown in Figure~\ref{fig:flownet}. The key idea is that we do not need to generate the dense flow maps from scratch since we can leverage the traditional feature matching algorithms like SURF~\cite{bay2006surf} to generate sparse corresponding points. The displacement of the corresponding points (Figure~\ref{fig:flownet} (b)) are considered as initial flow seeds and propagated to the entire region. In addition, the network is able to suppress the mismatches existing in the corresponding points during training. In this way,  plausible results can also be generated in the interior of non-texture regions.

As for the method of propagation, we adopt the architecture of CSPN proposed by Cheng et al.\cite{cheng2018depth,cheng2018depth2} for its efficiency. This method diffuses the information of the center pixel to its eight neighborhood iteratively in the form of convolution. Our adapted network is a very simple and common encoder-decoder architecture, which takes stacked RGB images and sparse seeds as input and outputs two results: one is the coarse optical flow $F_{0}$ and another is the  transformation kernels $\hat{K}_{i,j}$ with $k^2-1$ channels, where $k$ denotes the kernel size. Then the coarse optical flow is refined iteratively as in~\cite{cheng2018depth,cheng2018depth2}:

\begin{equation}
   F_{i,j,t+1} = \sum_{a,b=-(k-1)/2}^{(k-1)/2}K_{i,j}(a,b)\odot F_{i-a,j-b,t}
\end{equation}
where $\odot$ denotes element-wise product and:
\begin{equation}
K_{i,j}(a,b) = \frac{\hat{K}_{i,j}(a,b)}{\sum_{a,b\ne 0}|\hat{K}_{i,j}(a,b)|}
\end{equation}
\begin{equation}
K_{i,j}(0,0) = 1-\sum_{a,b\ne 0}K_{i,j}(a,b)
\end{equation}
Before each propagating operation, we fix the sparse seeds in order to guarantee that our propagated flows have the exact same value at those
valid pixels in the sparse flow map:

\begin{equation}
F_{i,j,t+1} = (1-m_{i,j})F_{i,j,t+1} + m_{i,j}F_{i,j}^{s}
\end{equation}
where $F_{i,j}^{s}$ denotes the sparse flow with empty positions filled with zero, $m_{i,j}$ is an indicator for the availability of sparse flow at $(i, j)$. In our training setting, the kernel size $k$ is set to 3 and the max steps of iterations are 16.


\subsection{PoseNet}

The PoseNet is also an important component of the unsupervised depth learning pipeline, which is an application of deep learning in visual odometry and responsible for estimating the pose in 6 degrees of freedom (DoF) between two images. In driving scenes like KITTI the poses are fairly simple and the cars in most images are just driving forward. But in indoor environments, the images are typically collected by handheld cameras, which means more complex ego-motion and raises the difficulty of learning for PoseNet.

There are also some approaches proposed to get rid of the PoseNet. For instance, Mahjourian et al.~\cite{mahjourian2018unsupervised} used Iterative Closest Point (ICP)~\cite{besl1992method,chen1992object,rusinkiewicz2001efficient} to compute a transformation that minimizes point-to-point distances between corresponding points. Wang et al.~\cite{wang2018learning} used direct visual odometry (DVO)~\cite{steinbrucker2011real} to obtain camera pose from predicted depth and images. But in our experiments these network-free methods collapsed during training on indoor datasets since they rely on good initial depth with low level of noise. Collapse means all the predictions of depth converge to a constant value. The estimation of poses can also be considered as a Perspective-n-Point (PNP) problem because we have predicted depth and dense matching simultaneously. We tried to implement the EPNP~\cite{lepetit2009epnp} algorithm on GPU to directly compute camera pose from predicted depth and flow in real time but the training collapsed again. These trials indicate that the use of PoseNet is necessary since the prediction of PoseNet is based on the entire dataset statistically and it is not likely to be dominated by individual training samples.

Therefore we rethink the working principle of the PoseNet. How does this blackbox estimate the pose from stacked RGB images? A reasonable speculation is that it first finds corresponding points in two images internally and infers the pose from the displacements of the corresponding points according to an unknown rule. However, there is no need for the PoseNet to match the pixels again since we already have the optical flow result, i.e., the dense matching. So we propose to use the flow result produced by SF-Net as the input of PoseNet instead of RGB images. 

This modification is equivalent to separate the unsupervised pose estimation into two stages: estimating the optical flow between two frames first and then inferring the pose based on the flow. It is unnecessary to do this separation if there exist ground-truth labels. But it enhances the interpretability of the PoseNet and reduces the difficulty of unsupervised learning. The significant improvement in the experimental results reported in Section~\ref{abla} also supports our speculation.

\subsection{Pure Rotation}
\label{sec:pure}
There still exists a problem that should not be ignored: pure rotation. This problem does not exist in driving scenes since previous methods generally remove static frames during preprocessing but is common in indoor datasets. Theoretically image pairs with pure rotation do not contain depth information and are harmful to the training. It is crucial to first filter out the training samples with pure rotation. It can be derived that the relationship between corresponding points with pure rotation can be fitted by a homography matrix $H$:

\begin{equation}
H=KRK^{-1}
\end{equation}
where $K$ denotes the camera intrinsic and $R$ denotes the rotation matrix. 

So for each image pair, we use the dense optical flow generated by SF-Net to compute its homography matrix with RANSAC~\cite{fischler1981random}. If the ratio of outliers is lower than a pre-set threshold (20\% in our setting, which means more than 80\% pixels can be fitted by a homography matrix), we consider the pose of this image pair as pure rotation and discard it. After this filtering process, about 30\% of the images in NYU V2 are discarded.

\subsection{Loss Function}

\subsubsection{Loss Function of SF-Net}

\quad For the training of SF-Net, we use photometric loss and smoothness loss.

\textbf{Photometric Loss}\quad This loss function computes the appearance difference between two images. We adopt the same setting as~\cite{godard2017unsupervised} which combines L1 loss and the Structural Similarity (SSIM)~\cite{wang2004image}. Besides per-pixel minimum trick proposed by~\cite{godard2018digging} is also adopted which is aimed to handle occlusion/disocclusion. It can be written as:

\begin{equation}
    L_{ph}= \sum_{p}\min_{s}(\alpha SSIM(I_t,I_{s\xrightarrow{}t})+(1-\alpha)|I_t(p) - I_{s\xrightarrow{}t}(p)|)
    \label{equ:lSSMI2}
\end{equation}
where $p$ indexes over pixel coordinates, $s$ denotes the index of source views, $\alpha$ is set to 0.5. 


\textbf{Smoothness Loss}\quad Although SF-Net adopts a sparse-to-dense training scheme, we also use edge-aware
flow smoothness loss to suppress the mismatches in initial seeds:

\begin{equation}
    L_{smooth}=\sum_{p}{\left|{\nabla}F(p)\right|\cdot \left(e^{-\left|{\nabla}I(p)\right|}\right)^T}
    \label{equ:lsmooth}
\end{equation}
where $\nabla$ is the vector differential operator, and $T$ denotes the transpose of image gradient weighting. So the total loss of SF-Net is:

\begin{equation}
    L=\lambda_{1}L_{ph}+\lambda_{2}L_{smooth}
\end{equation}

\begin{figure*}
\begin{center}
  \includegraphics[width=1\linewidth]{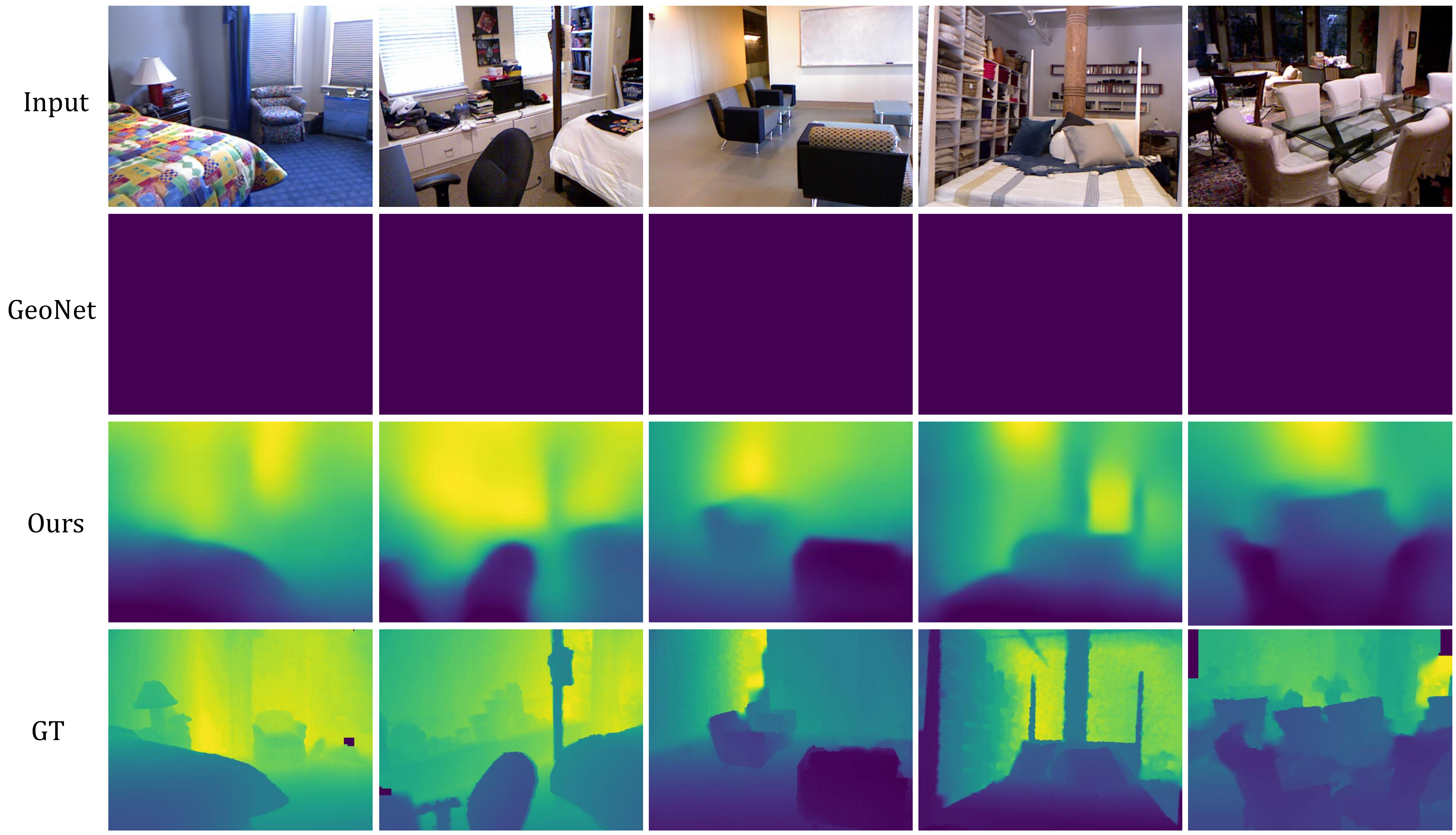}
\end{center}
   \vspace{-3mm}
   \caption{Qualitative comparison between GeoNet~\cite{yin2018geonet}, ours and ground-truth depth. We directly use the original code of GeoNet, which is able to achieve state-of-the-art performance on KITTI, but collapsed during training on NYU V2~\cite{silberman2012indoor}.}
   \vspace{-3mm}
\label{fig:nyu_depth}
\end{figure*}

\begin{table*}[htbp]
  \centering
    \begin{tabular}{ccccccccc}
    \toprule
    \multicolumn{2}{c}{\multirow{2}[4]{*}{Method}} & \multirow{2}[4]{*}{Supervision} & \multicolumn{3}{c}{Accuracy metric} & \multicolumn{3}{c}{Error metric} \\
\cmidrule(l){4-6} \cmidrule(l){7-9}    \multicolumn{2}{c}{} &       & $\delta<1.25$ & $\delta<1.25^2$ & $\delta<1.25^3$ & rel & log10 & rms \\
    \midrule
    \multicolumn{2}{c}{Make3D~\cite{saxena2009make3d}} & \checkmark  & 0.447 & 0.745 & 0.897&0.349 & - & 1.214 \\
    \multicolumn{2}{c}{Depth Transfer~\cite{karsch2014depth}} & \checkmark & - & - & - & 0.35 & 0.131 & 1.2 \\
    \multicolumn{2}{c}{Liu et al.~\cite{liu2014discrete}} & \checkmark & - & - & - & 0.335 & 0.127 & 1.06  \\
    \multicolumn{2}{c}{Ladicky et al.~\cite{ladicky2014pulling}} &\checkmark & 0.542 & 0.829 & 0.941 & - & - & -  \\
    \multicolumn{2}{c}{Li et al.~\cite{li2015depth}} & \checkmark & 0.621 & 0.886 & 0.968 & 0.232 & 0.094 & 0.821  \\
    \multicolumn{2}{c}{Wang et al.~\cite{wang2015towards}} & \checkmark & 0.605 & 0.890 & 0.970 & 0.220 & - & 0.824  \\
    \multicolumn{2}{c}{Roy et al.~\cite{roy2016monocular}} & \checkmark & - & - & - & 0.187 & - & 0.744  \\
    \multicolumn{2}{c}{Liu et al.~\cite{liu2016learning}} & \checkmark & 0.650 & 0.906 & 0.976 & 0.213 & 0.087 & 0.759  \\
    \multicolumn{2}{c}{Li et al.~\cite{li2017two}} & \checkmark & 0.788 & 0.958 & 0.991 & 0.143 & 0.063 & 0.635  \\
    \multicolumn{2}{c}{MS-CRF~\cite{xu2017multi}} & \checkmark & 0.811 & 0.954 & 0.987 & 0.121 & 0.052 & 0.586  \\
    \multicolumn{2}{c}{DORN~\cite{fu2018deep}} & \checkmark & 0.828 & 0.965 & 0.992 & 0.115 & 0.051 & 0.509  \\
    \midrule
    \multicolumn{2}{c}{Ours (baseline$\dagger$)} & $\times$ & 0.511 & 0.779 & 0.904 & 0.331 & 0.127 & 1.000 \\
    \multicolumn{2}{c}{Ours} & $\times$ & 0.674 & 0.900 & 0.968 & 0.208 & 0.086 & 0.712 \\
    \bottomrule
    \end{tabular}%
  \vspace{2mm}
  \caption{\label{tab:nyu} Comparison to existing methods on NYU V2~\cite{silberman2012indoor}. $\dagger$ denotes the model that collapsed during training. Since we adopt the scale normalization proposed by~\cite{wang2018learning} to avoid the shrinking of depth, all the predictions are normalized to 1 meter when the model collapses. All the other methods in the table are fully supervised by depth annotations.}
\end{table*}%

\begin{table*}[htbp]
\renewcommand\tabcolsep{4pt}
  \centering
    \begin{tabular}{ccccccccc}
    \toprule
    \multirow{2}[4]{*}{Method} & \multirow{2}[4]{*}{Supervision} & Filtering of & \multicolumn{3}{c}{Accuracy metric} & \multicolumn{3}{c}{Error metric} \\
\cmidrule(l){4-6} \cmidrule(l){7-9}          &       & Pure Rotation & d1 & d2 & d3 & rel   & log10 & rms \\
    \midrule
    DepthNet + R-PoseNet $\dagger$  & RGB   & $\times$    &  0.511 & 0.779 & 0.904 & 0.331 & 0.127 & 1.000 \\
    DepthNet + R-PoseNet $\dagger$  & RGB   & \checkmark   &  0.511 & 0.779 & 0.904 & 0.331 & 0.127 & 1.000 \\
    DepthNet + F-PoseNet $\dagger$  & Flow (w/o propagation)& $\times$   &  0.511 & 0.779 & 0.904 & 0.331 & 0.127 & 1.000 \\
    DepthNet + F-PoseNet  & Flow (w/o propagation) & \checkmark   & 0.596 &  0.862  &  0.951  &  0.257  &  0.102 & 0.841 \\
    DepthNet + R-PoseNet  & Flow (w/ propagation) & \checkmark   & 0.578  & 0.836  & 0.938 & 0.273 & 0.108 & 0.910 \\
    DepthNet + F-PoseNet  & Flow (w/ propagation) & \checkmark   & 0.674 & 0.900 & 0.968  & 0.208 & 0.086 & 0.712 \\
    \bottomrule
    \end{tabular}%
  \vspace{2mm}
  \caption{\label{tab:ablation} Evaluation of each component on NYU V2's test split. $\dagger$ denotes the model that collapsed during training. R-PoseNet denotes the PoseNet with RGB images input and F-PoseNet denotes the PoseNet with flow input. $\delta<1.25, \delta<1.25^2, \delta<1.25^3$ are abbreviated to d1, d2, d3 due to space limitation.}
  \vspace{-2mm}
\end{table*}%

\begin{table}[htbp]
  \centering
  \renewcommand\tabcolsep{1.8pt}
  \newcommand{\tabincell}[2]{\begin{tabular}{@{}#1@{}}#2\end{tabular}}
  \resizebox{1.0\linewidth}{!}{
    \begin{tabular}{cccccc}
    \toprule
    \multirow{2}[4]{*}{Method} & \multirow{2}[4]{*}{Dataset} & Accuracy&\multicolumn{3}{c}{Error} \\
\cmidrule(l){3-3} \cmidrule(l){4-6}         &   & d1 & rel   & sq rel & rms  \\
    \midrule
    Ours(baseline)$\dagger$ & S     & 0.631& 0.238 &  0.190 & 0.570  \\
    Ours(F-PoseNet)$\dagger$ & S& 0.631 & 0.238 &  0.190 & 0.570  \\
    Ours(F-Sup+R-PoseNet) & S     & 0.682& 0.206 &  0.134     & 0.491  \\
    Ours(F-Sup+F-PoseNet) & S     & {\bf 0.710} &{\bf 0.190}  & {\bf 0.124}  & {\bf 0.465} \\
    \midrule
    Ours(baseline) & K     & 0.833 & 0.132 & 1.016  & 5.506  \\
    Ours(F-Sup) & K     & {\bf 0.837} & 0.136  & 1.110 &  5.327  \\
    Ours(F-PoseNet) & K     & 0.836  & {\bf 0.130} & {\bf 1.001} &  {\bf 5.294}    \\
    \bottomrule
    \end{tabular}%
    }
  \label{tab:scannet1}%
  \vspace{-1mm}
  \caption{Evaluation of depth on Scannet and KITTI datasets. K:KITTI, S:Scannet, $\dagger$:collapse, F-sup:flow supervision.}
\end{table}%

\subsubsection{Loss Function of DepthNet and PoseNet}

\quad For the training of DepthNet and PoseNet, the loss function consists of four terms.

\begin{figure}[htbp]
\vspace{-3mm}
\begin{center}
  \includegraphics[width=1\linewidth]{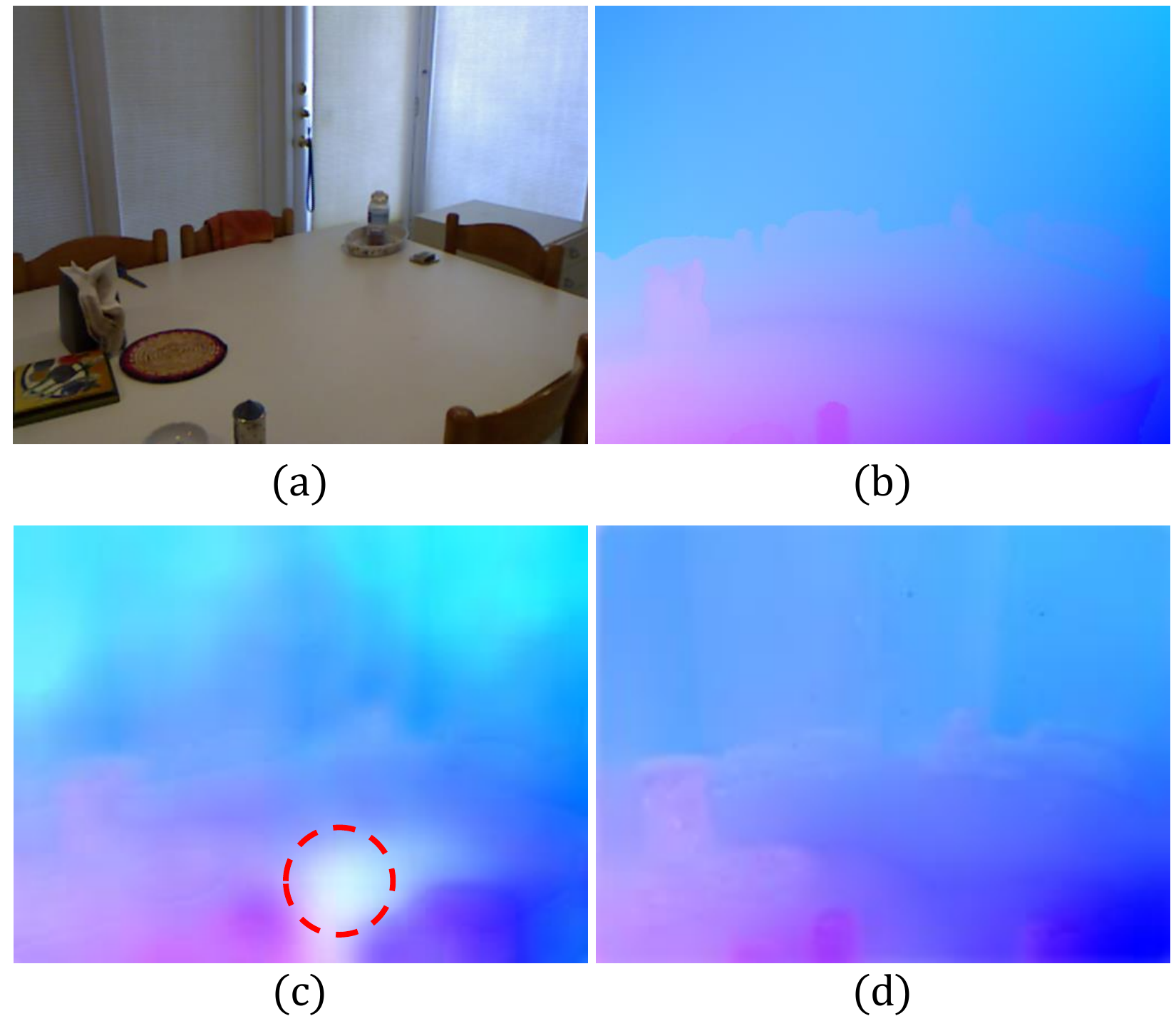}
\end{center}
   \vspace{-3mm}
   \caption{(a) A typical image with large areas of non-texture regions. (b) GT flow. (c) Flow without propagation. (d) Flow with propagation. Previous unsupervised flow learning methods only impose smoothness constraint on non-texture regions, which are prone to consider these areas as static (red circle in (c)).}
\label{fig:compare_flow}
\vspace{-3mm}
\end{figure}

\textbf{Rigid Flow Loss}\quad This term directly uses the optical flow result produced by SF-Net as supervision of the synthsized rigid flow. We use berHu~\cite{laina2016deeper} norm $||\cdot||_{\delta}$ to measure the deviation:

\begin{equation}
    L_{flow} = ||f_{t\xrightarrow{}s}(p_t) - f'_{t\xrightarrow{}s}(p_t)||_{\delta}
\end{equation}

The other two terms $L_{s\_depth}$ and $L_{s\_normal}$ are smoothness constraints and are similar to Equation~\ref{equ:lsmooth}. But they are imposed on the predicted depth and corresponding normal which is computed from the depth. The last term $L_{ph}$ is the same as Equation~\ref{equ:lSSMI2}. So the total loss of DepthNet and PoseNet is
\begin{equation}
    L=\lambda_{1}L_{flow} + \lambda_{2}L_{ph}+\lambda_{3}L_{s\_depth}+\lambda_{4}L_{s\_normal}
\end{equation}

\begin{table}[htbp]
\vspace{-3mm}
\renewcommand\tabcolsep{4pt}
  \centering
    \begin{tabular}{cccc}
    \toprule
    Method & w/o Propagation & w/ Propagation  & PWC-Net \\
    \midrule
    EPE   & 7.409 & 3.602  & 3.279 \\
    \bottomrule
    \end{tabular}%
  \vspace{2mm}
  \caption{\label{tab:sa_region} Average endpoint error (EPE) flow results on NYU validation split. The result of PWC-Net is directly generated by PWC-Net~\cite{Sun2018PWC-Net} pretrained on FlyingChairs~\cite{DFIB15} with supervision but without finetuning on NYU V2. The result obtained by SF-Net with propagation is close to supervised result obtained by PWC-Net.}
  \vspace{-2mm}
\end{table}%

\begin{figure*}
\begin{center}
  \includegraphics[width=1\linewidth]{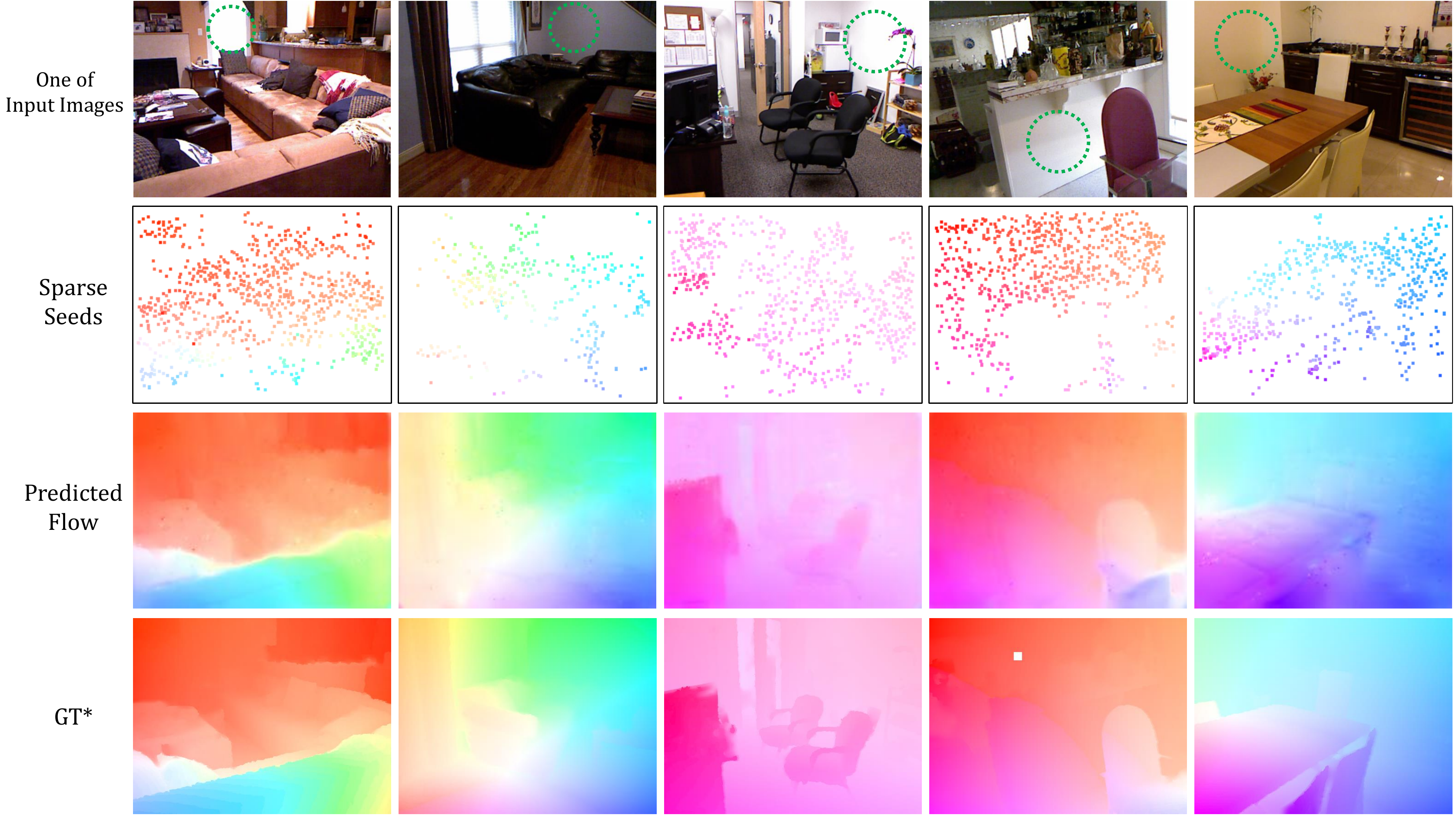}
\end{center}
   \vspace{-3mm}
   \caption{Visualization of the results generated by SF-Net on NYU V2. NYU depth dataset does not contain optical flow annotations. * means that the flow results are computed with ground-truth depth by the method mentioned in Section~\ref{abla}. Our SF-Net handles the non-texture regions (green circles) well. The sizes of sparse points have been enlarged for better visualization.}
\label{fig:kitti_quqa}
\vspace{-3mm}
\end{figure*}

\section{Experiments}

\subsection{Comparisons on NYU V2 Dataset}

The NYU Depth v2~\cite{silberman2012indoor} dataset contains 582 indoor video scenes taken with a Microsoft Kinect camera and the training split contains 283 scenes (about 230K images). To train the DepthNet and PoseNet, we first use the method mentioned in Section~\ref{sec:pure} to filter out the image pairs with pure rotation. About 30\% images are discarded and finally we use about 180K images for training. We fix the length of training image sequences to be 3 frames for all the three networks, and treat the central frame as the target view and the $\pm 10$ frames as the source views. We only use raw RGB image sequences for training and the images are resized to $192\times 256$. Visualization is shown in Figure~\ref{fig:nyu_depth} and quantitative evaluation is reported in Table~\ref{tab:nyu}. The baseline model only consists of DepthNet and PoseNet with RGB images input and does not use flow as supervision.


\subsection{Ablation Study}

\label{abla}

In this subsection, we individually evaluate the effects of the four components in our pipeline: 1) The flow supervision. 2) The filtering of pure rotation.  3) The propagation design for the SF-Net. 4) The input of the PoseNet. 

As shown in Table~\ref{tab:ablation}, the results of rows 3,6 and rows 4,5 indicate that the flow supervision and filtering of pure rotation are indispensable in order to get rid of the collapse phenomenon. It also shows that both the F-PoseNet and the propagation of SF-Net improve the performance significantly. The result without propagation means the SF-Net takes RGB images as input and only output optical flow, and the other parts in the network remains the same.

NYU depth dataset does not contain optical flow annotations. But we can compute the rigid flow from the ground-truth depth. We first use the EPNP~\cite{lepetit2009epnp} algorithm to solve the pose between two images with depth annotations and sparse matchings. The sparse matchings are also obtained by SURF~\cite{bay2006surf}. Then the estimated pose and ground-truth depth can be used to compose the rigid flow. We consider these estimated rigid flows as ground-truth flow labels.  

In order to quantificationally evaluate the performance of SF-Net, we randomly choose 1000 image pairs as validation set and exclude them during training. The quantitative result is shown in Table~\ref{tab:sa_region} and qualitative comparison is shown in Figure~\ref{fig:compare_flow}. With the help of the propagation of sparse seeds, SF-Net handles the non-texture regions well and achieves reasonably good results as shown in Figure~\ref{fig:kitti_quqa}.

\vspace{-1mm}
\subsection{Evaluations on Scannet and KITTI}

As shown in Table 3, we also evaluate our method on Scannet and KITTI datasets. Since Scannet is a large dataset, we train and test our model on a subset of Scannet for efficiency, which contains 40 scenes and about 70K images. The flow supervision is proposed to address the non texture problems in indoor scenarios, but has its own limitation. When applied on KITTI, the accuracy of the flow becomes the bottleneck.  However, F-PoseNet still has gain.

\section{Conclusion}

In this paper, we propose a new unsupervised depth learning framework which reduces the difficulty for the network to learn and is able to work indoor. We also propose a sparse-to-dense unsupervised flow estimation network that addresses the intractable non-texture area problem. More importantly, the results of our approach demonstrate that the technology of unsupervised depth learning is not only able to work in driving scenes but also has the capacity to be applied in more general scenes. This is an important step towards exploring the innumerable videos available on the Internet for the training of deep learning.

\textbf{Limitation}\quad Although our approach is able to handle the non-texture regions in most cases, it relies on the propagation of sparse seeds. In some special situations where non-texture areas are extremely large and the identified corresponding key points are very sparse, it is difficult for our model to predict correct results. In addition, the generated flows usually have blurry boundary. When these blurry flows are used to supervise the training of DepthNet, the obtained depth maps have blurrier boundary. This also limits the performance of our model.
We will address these problems in the future work.
{\small
\bibliographystyle{ieee_fullname}
\bibliography{egbib}
}

\end{document}